\documentclass[preprint,12pt,3p]{elsarticle}

\usepackage{amssymb}

\usepackage{comment}
\usepackage{url}
\usepackage{hyperref}
\usepackage{longtable}
\usepackage{graphicx}
\usepackage{subcaption}
\usepackage{color,soul}
\usepackage[]{algorithm2e}

\journal{Online Social Networks and Media}

\begin{document}

\begin{frontmatter}

\title{Who framed Roger Reindeer? \\
De-censorship of Facebook posts by snippet classification}


\author[label1]{Fabio Del Vigna}
\address[label1]{Institute of Informatics and Telematics (IIT-CNR), via G. Moruzzi 1, Pisa, Italy\fnref{label4}}

\cortext[cor1]{Corresponding author}

\ead{f.delvigna@iit.cnr.it}

\author[label1]{Marinella Petrocchi\corref{cor1}}
\ead{m.petrocchi@iit.cnr.it}

\author[label2]{Alessandro Tommasi}
\address[label2]{LUCENSE SCaRL, Traversa prima di Via della Chiesa di Sorbano del Giudice n. 231, Lucca, Italy\fnref{label3}}
\ead{alessandro.tommasi@lucense.it}

\author[label2]{Cesare Zavattari}
\ead{cesare.zavattari@lucense.it}

\author[label1]{Maurizio Tesconi}
\ead{m.tesconi@iit.cnr.it}

\begin{abstract}
This paper considers online news censorship and it concentrates on censorship of identities. Obfuscating identities may occur for disparate reasons, from military to judiciary ones. In the majority of cases, this happens to protect individuals from being identified and persecuted by  hostile people.
However, 
being the collaborative web characterised by a redundancy of information, it is not unusual that the same fact is reported by multiple sources, which may not apply the same restriction policies in terms of censorship. Also, the proven aptitude of social network users to disclose personal information leads to the phenomenon that comments to news can reveal the data withheld in the news itself. 
This gives us a mean to figure out who the subject of the censored news is. We propose an adaptation of a text analysis approach to unveil censored identities. The approach is tested on a synthesised scenario, which however resembles a real use case. 
Leveraging a text analysis based on a context classifier trained over snippets from posts and comments of Facebook pages, we achieve promising results. Despite the quite constrained settings in which we operate -- such as considering only snippets of very short length -- our system successfully detects the censored name, choosing among 10 different candidate names, in more than 50\% of the investigated cases. This outperforms the results of two reference baselines. The findings reported in this paper, other than being supported by a thorough experimental methodology
and interesting on their own, also pave the way for further investigation on the insidious issues of censorship on the web.\end{abstract}

\begin{keyword}

News Censorship \sep Identities Censorship \sep Text Mining \sep Data Correlation \sep Privacy in Social Networks \sep Social Media Mining \sep  Facebook \sep Named Entity Recognition \sep Candidate Entity Recognition \sep Transparency
\end{keyword}

\end{frontmatter}





\section{Introduction}\label{sec:intro}

With billions of users, social media probably represent  the most privileged channel for publishing, sharing, and commenting information. In particular, social networks are often adopted to spread news content~\cite{Lerman2010}. According to a Pew Research study, Americans often gets news online, with a double share of them preferring social media rather than print magazines\footnote{The Modern News Consumers, a Pew Research study: \url{http://www.journalism.org/2016/07/07/pathways-to-news/}, July, 7, 2016 -- ; All URLs in this paper have been lastly accessed on February, 20, 2018.}. As a matter of fact, popular newspapers have an official account on social platforms. Through their pages, news stories - or their previews - are published - often under the form of a short post, with a further link to the complete text. The readers' community can like, share, and re-post news stories. Users can also comment and discuss issues in the news themselves. Still a Pew Research survey highlights that a share of 37\% of social media news consumers comment on news stories, while the 31\% ``discuss news on the news in the site"\footnote{10 facts about the changing of the digital landscape: \url{http://www.pewresearch.org/fact-tank/2016/09/14/facts-about-the-changing-digital-news-landscape/}, September 14, 2016.}. 
Undeniably, users' comments and discussions may help to increase the awareness and value of the published information, thanks to the addition of informative details. Examples of support are, for example, the depiction of the context in which the news facts took place, or to track down mistakes, draw rectifications, and even unveil fake information. 

In this paper, we focus on online news articles and, in particular, on those news portions that organisations choose not to make public. 
News censorship may occur for different reasons. Organisations, be them military, commercial, governmental, or judicial, may decide to veil part of the information to protect sensitive data from, e.g., competitors, customers, or hostile entities. Standard examples of censored data are identities: from a business point of view, a press agency may veil the identity of the buyer of a huge amount of fighters. Also, the names of the victims of particularly hateful offences, like rapes and abuses on minors, are typically obfuscated, as for regulations dictated by law. Finally, a peculiar practice when publishing Israeli military-related news on social media is the veiling  of the identities of public officers (e.g., Corporal S., rather than the explicit identity of such officer, see, e.g.,~\cite{Cascavilla2015}). 
However, as highlighted by recent literature~\cite{Schwartz:2017}, given the essential nature of social networking, the ``non identification alone is ineffective in protecting sensitive information". This is due to the fact that, featuring a {\it commented post} structure of the published news, a specific information, withheld in the news, is compromised through the effects of users' comments, where specific content may reveal, either {\it explicitly} or {\it implicitly}, that information.


This work places itself amongst a few ones, like, e.g.,~\cite{Burattin2014,Cascavilla2015} that investigate to which extent the connections among news articles, comments, and social media influence the effectiveness of identities censorship procedures. In particular, we present a novel approach to unveil a censored identity in a news post, by exploiting the fact that, on the social Web, it is not unusual to find the same content, or a very similar one, published elsewhere, e.g., by another publisher with different censorship policies. Also and noticeably, as discussed above,  the amount of user generated content on social networks may lead to the very unexpected phenomenon according to which the hidden information may emerge in the users' comments.

 Differently from prior work in the area, which exploits  the friendship network of the commenters to some censored news, here we inherit from the field of text analysis.
 In particular, we exploit techniques often used to address co-reference resolution~\cite{Clark08coreferenceresolution:}, based on recognising the context in which certain names tend to appear, to successfully address the task of unveiling censored names. To the best of our knowledge, this is  the first attempt that addresses the task by means of a  semi-supervised approach, which only makes use of texts, without relying on metadata about the commenters, and trying to reconstruct missing information  exploiting  similar contexts. For running and validating our analysis, we make use of Facebook data, which we {\it purposely} censored for the sake of the experiments. Even if we rely on an experimental setting that is built {\it ad hoc}, our synthesised scenario is easily connectable to real use cases, as described in subsequent sections.
 
 Our extensive experimental campaign explores a dataset of almost 40,000 posts published on the Facebook pages of the top 25 US newspapers (by weekday circulation). 
By exploiting an algorithm based on context categorisation, we train a classifier on the posts, and related comments, in the dataset, to demonstrate the capability to reveal the censored term. 
The system performances are benchmarked against two baselines, obtaining a more than significant improvement. 

Summarising, the paper contributes along the following dimensions:
\begin{itemize}
    \item the design and development of a methodology based on text-analysis, here applied for the first time to spot identities that have been censored in social media content;
     \item the proposal of an approach that is solely based on very loosely structured data, in contrast to other proposed techniques that leverage the social network structure. The latter have the issues that 1.\ the association between names and social network nodes needs to be addressed, and 2.\ the structure of the social network constitutes significant a-priori knowledge. Instead, we simply use raw data, by only assuming a ``commented post" structure of the data;
    \item starting from revealing censored popular identities, our results constitute the prelude to the detection of other kind of censored terms, such as, e.g., brands and even identities of common people, whose veiling is a usual practice often applied by publishers for privacy issues, be them driven by legal, military, or business motivations. 
\end{itemize}





 In the next section,   we first introduce real identity censorship procedures, discussing the role of comments - and commenters - in bypassing their effectiveness. Section~\ref{sec:datasets} presents
the data corpus for our analyses, also highlighting similarities of such corpus with the real scenarios presented in Section~\ref{sec:scenarios}. Section~\ref{sec:methodology} presents the methodology, and Section~\ref{sec:exp-res} describes the experiments and comments the results. In Section~\ref{sec:relwork}, we discuss related work in the area of investigation. 
Finally, Section~\ref{sec:concl} concludes the paper.


\section{Identities censorship in online news and its circumvention}\label{sec:scenarios}
Moving from motivations for identities censorship in news, this section discusses the effectiveness of such censorship when news are published online~\cite{Schwartz:2017}.





Traditionally, the censorship of identities in news occurs for three main reasons: 1) business, since, e.g., it may be not advantageous to disclose the real identity of a participant in a commercial transaction, such as a large quantities of weapons; 2) legal (e.g., do not become known minors abused); and 3) military, to protect the individuals, and their relatives, from being identified by adversaries. As an example,  in Israeli `` policy dictates many situations in which the identity of officers must not be released to the public~\cite{Schwartz:2017}. In the above circumstances, the censorship usually takes place either by putting an initial or using a fancy name. 


With the advent of the social media era, the publication of news on social networks sites became a usual pratice. Thus, when the news is published on social networks sites, such as on the Facebook pages of newspapers, the identities are still blurred as written above, directly from the organisation that chooses not to publish that information (therefore, either the government, or the news agency that publishes the news, or some other military or commercial stakeholder). 

However, when the news post is on the social network, and a ``commented post" structure of the data is followed, the comments are freely posted by users other than the news publisher. Also, comments are generally not moderated by the platform administrators, unless they are reported as offensive content, inciting hate campaigns, or some judicial authority required the cancellation of specific comments.


The fact that there are a number of uncensored comments leads to the phenomenon that, although in the post the information is withheld, that information is compromised by one, or more, comments. In fact, it has been  proven that, although the organisation that posted the news has censored an identity in the news itself, and so published it, those people who know the censored name and who make comments tend to talk about the name, indirectly or even directly. This is the case featured, e.g., by the Facebook dataset analysed in~\cite{Schwartz:2017}, where 325,527 press items from 37 Facebook news organisation pages were collected. A total of 48 censored articles were identified by a pattern matching algorithm first, and then manually checked. On the whole amount of comments tied to those articles, the 19\% of them were classified as comments presenting an explicit identification of the name or the use of a pseudonym. A de-censorship analysis based on the social graph of the commenters has been carried out to recognise the censored names~\cite{Cascavilla2015}. In the rest of the paper, we will propose a methodology based instead of recognising the textual context in which certain terms tend to appear.


To test the methodology, we rely on a Facebook dataset that we intentionally censored, by however resembling the real scenarios depicted above. We  deliberately censored identities in a Facebook dataset of US newspaper posts, leaving comments to posts unchanged. The dataset is described in the following section.


\section{Dataset of US newspapers Facebook pages}\label{sec:datasets}

In this work, we leverage the fact that many newspapers have a public social profile, to check whether an identity - that was censored in some news - can be spotted by analysing comments to that news - as well as other news published on different online newspapers. 
We consider a set of posts and comments
from the Facebook pages of the top 25 newspapers, by weekday circulation, in US\footnote{\url{https://en.wikipedia.org/wiki/List_of_newspapers_in_the_United_States}}. Table~\ref{tab:dataset} shows the names of the newspapers, the corresponding Facebook page (if any), and the number of collected posts and comments, crawled from the pages.
\begin{figure}
    \centering
    \includegraphics[width=8cm]{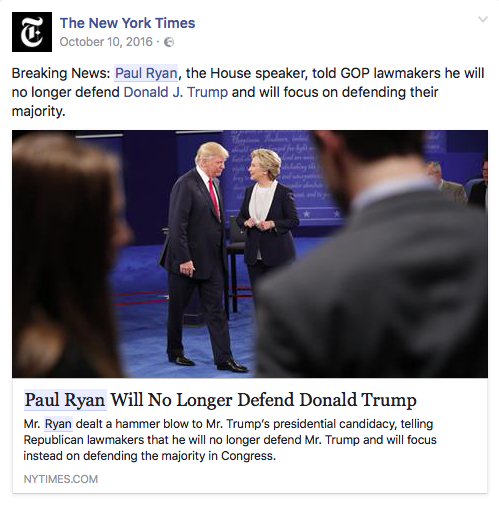}
    \caption{Example of a post published on the NYT Facebook page.}
    \label{fig:post}
\end{figure}

We developed a specific crawler to collect the data. The crawler is written in the PHP scripting language, using the Laravel\footnote{\url{https://laravel.com}} framework, to make it scalable and easy manageable. In particular, the crawler exploits the Facebook Graph API\footnote{\url{https://developers.facebook.com/docs/graph-api}} to collect all the public posts, comments, and comments to comments from the Facebook pages of the newspapers.
The collection requires as input the URL of the Facebook page and a set of tokens required to authenticate the application on the social network. The crawler supports parallel downloads, thanks to its multi-process architecture. It recursively downloads data, until it fully covers the time span specified by the operator. It stores data in the JSON format, since it is the most natural format for social media data. Also, such format can  be easily employed to feed storage systems like Elasticsearch\footnote{\url{https://www.elastic.co}} or MongoDB\footnote{\url{https://www.mongodb.com/it}}.

We collected posts and comments from August 1, 2016 to October 31, 2016. Overall, we obtained 39,817 posts and 2,125,038 comments.

\begin{table}[!ht]
\begin{small}
	\begin{centering}
		\begin{tabular}{|l|l|c|c|}
		\hline
	 	\bf{Newspaper} & \bf{Facebook profile} & \bf{Posts} & \bf{Comments}\\
		\hline
		The Wall Street Journal & wsj & 1577 & 136969 \\
		\hline
		The New York Times & nytimes & 265 & 98099 \\
		\hline
		USA Today & usatoday & 560 & 155893 \\
		\hline
		Los Angeles Times & latimes & 532 & 124477 \\
		\hline
		San Jose Mercury News & mercurynews & 0 & 0\\
		\hline
		New York Daily News & NYDailyNews & 1637 & 124948 \\
		\hline
		New York Post & NYPost & 479 & 132715 \\
		\hline
		The Washington Post & washingtonpost & 232 & 101260 \\
		\hline
		Chicago Sun-Times & thechicagosuntimes & 2215 & 64675 \\
		\hline
		The Denver Post & denverpost & 1376 & 113621 \\
	    \hline
		Chicago Tribune & chicagotribune & 2401 & 141361 \\
		\hline
		The Dallas Morning News & dallasmorningnews & 2458 & 148154 \\
		\hline
		Newsday & newsday & 2432 & 60549 \\
		\hline
		Houston Chronicle & houstonchronicle & 1350 & 920 \\
		\hline
		Orange County Register & ocregister & 1123 & 37153 \\
		\hline
		The Star-Ledger & Star.Ledger & 284 & 3142 \\
		\hline
		Tampa Bay Times & tampabaycom & 1539 & 76388 \\
		\hline
		The Plain Dealer & ThePlainDealerCLE & 4 & 33 \\
		\hline
		The Philadelphia Inquirer & phillyinquirer & 2124 & 10491 \\
		\hline
		Star Tribune & startribune & 2820 & 106357 \\
		\hline
		The Arizona Republic & azcentral & 2073 & 151590 \\
		\hline
		Honolulu Star-Advertiser & staradvertiser & 3487 & 52447 \\
		\hline
		Las Vegas Review-Journal & reviewjournal & 3588 & 108614\\
		\hline
		San Diego Union-Tribune & SanDiegoUnionTribune & 2163 & 45530 \\
		\hline
		The Boston Globe & globe & 3098 & 129652 \\
		\hline
		\end{tabular}
		\caption{Top 25 US newspapers by weekday circulation (as of March, 2013).}
		\label{tab:dataset}
	\end{centering}
\end{small}
\end{table}


\begin{figure}[!ht]
    \centering
    \includegraphics[width=8cm]{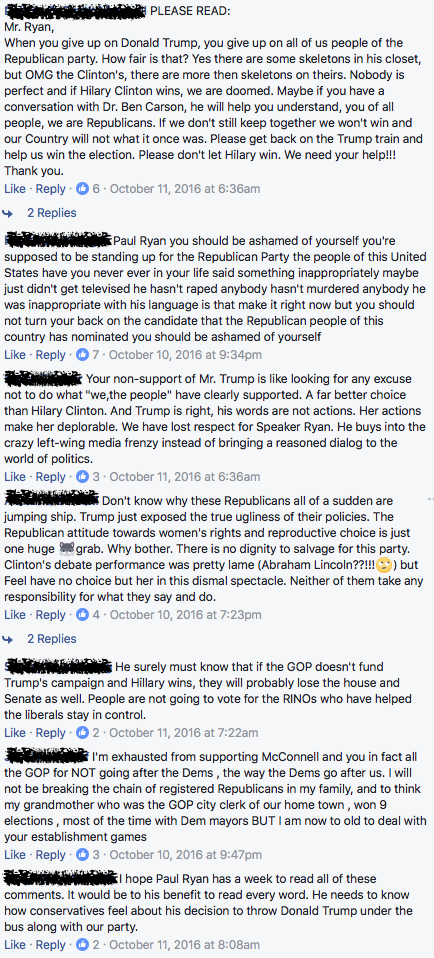}
    \caption{Excerpt of comments tied to the NYT post in Fig.~\ref{fig:post}.}
    \label{fig:comments}
\end{figure}

Figure~\ref{fig:post} reports an example of a typical post, published on the Facebook page of one of the most influential American newspapers, The New York Times. As shown in the figure \ref{fig:post}, the text of the post is short and not very informative, since the actual content of the news is in the link to the journal website. In that very short piece of text, two identities are mentioned, Paul Ryan and Donald J. Trump. Figure~\ref{fig:comments} shows an example list of some comments related to the same post. Comments  are usually copious, with discussions that are focused on several aspects. Notably, not all the comments are strictly related to the topics of the post. As we can see from the figure, more than one identity is mentioned in the comments set (i.e., Donald Trump, Hillary Clinton, Ben Carson, Paul Ryan, Abraham Lincoln, and McConnell) among which the two mentioned in the related post. Also, references to some of the identities are with different variants (e.g., Ryan, Paul Ryan, Trump, Donald Trump).  

It is  worth noting that comments are generally short, thus comparable to microblog messages, e.g., tweets or Tumblr posts. 

For the sake of the experiments illustrated in the next section, we purposely censored the identities of some people in part of the crawled Facebook posts,  to simulate a censorship behaviour. 

In the following, we will propose a methodology to recognise the censored name among a set of candidate names as they appear in comments to the censored post. Among the candidate names, there is the same name as the one in the censored post (e.g., Trump in the censored post, Trump in the comments). Remarkably, since our analysis is based on similarities of the {\it surroundings} of a name, rather than on the name itself, the proposed approach is still valid also when the censored identity is referred in comments with a pseudonym (e.g., Trump in the censored post, Mr. President in the comments). This last case often happens in real scenarios, as introduced in Section~\ref{sec:scenarios}. 


\section{Methodology}\label{sec:methodology}
The text analysis approach proposed in this paper to unveil censored identities  is motivated by  the hypothesis that, in the collaborative web, a ``perfect" censorship is impossible. Indeed,  the redundancy of information (e.g., the same fact reported by multiple sources, many of which do not apply the same restriction policy for publishing that fact) gives us a means to figure out who the subject of a censored post is.
Obviously, as shown in the post and comments of Figures~\ref{fig:post} and~\ref{fig:comments}, the same news reported and/or commented by various sources will not use the same exact wording (see, for example, Donald J. Trump, Donald Trump, Trump), so that the problem remains a non trivial one. 


\subsection{Overall methodology}\label{subsec:meth}
The proposed approach  makes use of two distinct Named Entity Recognisers (NER). Named Entity Recognition (NER) is the process of identifying and classifying
entities within a text. Common
entities to identify are, e.g., persons, locations, organizations, dates, times, and so on.  NER state-of-the-art systems
use statistical models (i.e., machine learning) and typically require a set of manually annotated training data, in combination with a classifier.
Popular NERs frequently adopted in the literature
are the 
Stanford NER tagger\footnote{\url{https://nlp.stanford.edu/software/CRF-NER.shtml}}, also available through the NLTK Python library\footnote{\url{https://www.nltk.org}}, and the spaCy NER (the one adopted in this paper, introduced in Section~\ref{subsec:selnames}).

Mostly based on machine learning techniques, NERs exploit features such as strong indicators for names (e.g., titles like ``Ph.D.'', ``Mr.'', ``Dr.'', etc.) to determine whether a small chunk of text  (say, a window of 50-200 characters around the name) indicates a person, an organisation, a date, and so on. NERs are to this day standard pieces of an NLP pipeline; we plan here on showing how to build on top of one to address our specific task. We first apply a generic NER to the whole collection of data depicted in Table~\ref{tab:dataset}, to detect the identities within. Then, by extending the NER scope to a wider context, we exploit a second, more specific entity recogniser, hereafter called Candidate Entity Recogniser (CER),  to recognise specific people (out of a smallish set of candidates).

Let's suppose that, in a piece of text (like a Facebook post) the name   ``Mark Smith'' has been censored by replacing it with ``John Doe''. The text might contain sentences such as: ``during his career as a drug dealer, John Doe was often in contact with the Medell\'in Cartel''. In order to reveal the original name hidden by John Doe, we will proceed as follows:  
\begin{enumerate}
    \item in the text, we will identify names to ``resolve'' (John Doe in the example), using the generic NER; 
    \item still relying on the generic NER, we will then identify a set of candidate names for the actual identity of John Doe, among which the correct one (``Mark Smith''), by searching in a textual data collection wider than the single piece of censored text.  
    For the sake of this simple example, let us assume that we have ``Mark Smith'' and ``Mary Jones'' as possible candidates;
    \item we will obtain a set of sentences that include the candidates, e.g., 
    ``The notorious drug baron Mark Smith used to spend much time in Medell\'in'' and ``Mary Jones's career as police officer was characterised by her fights with the drug dealers'';
    \item using the sentences retrieved at the previous step, we will train a customised version of a NER, i.e., the Candidate Entity Recogniser (CER), to identify instances of Mark Smith and Mary Jones, discarding the actual name from the text.
    In practice, we will train the CER with sentences like:
    ``The notorious drug baron {\textless}MarkSmith{\textgreater} XXXX XXXX {\textless}/MarkSmith{\textgreater} used to spend much time in Medell\'in'' or ``{\textless}MaryJones{\textgreater} XXXX XXXX {\textless}/MaryJones{\textgreater}'s career as police officer was characterized by her fights with the drug dealers'';
    \item finally, we will apply the CER model thus obtained to the censored sentences in the original text, to find out who they refer to. In practice, by feed the sentence: ``during his career as a drug dealer, XXXX XXXX was often in contact with the Medell\'in Cartel'', we expect the trained CER to return ``Mark Smith'' rather than ``Mary Jones".
\end{enumerate}

For the sake of a more compact representation, we synthesise the algorithm as follows:\\

\begin{algorithm}[H]
 \KwData{$dataset$, $target$}
 \KwResult{$candidate$}
 $identities \leftarrow \emptyset$\;
 $namedEntities \leftarrow \emptyset $\;
 $contexts \leftarrow \emptyset$\;
 $CER \leftarrow \emptyset$\;
 \For{$text \in dataset$} {
  Find all names in $text$ with the NER and put them in $namedEntities$\;
  \For {each $namedEntity$ found}{
   Put all sentences from $text$ that contains the $namedEntity$ in $contexts[namedEntity]$, substituting $namedEntity$ with XXXX XXXX\;
  }
 }
 \For {$namedEntity \in namedEntities$}{
  Train $CER[namedEntity]$ using $contexts[namedEntity]$\;
 }
 Apply $CERs$ to the $target$ and check which named entity better fits.

 \caption{How to train a \textit{CER} to spot censored identities.}
\end{algorithm}
\medskip

It is worth noting that the application of the algorithm to the whole collection would be costly, due to the large amount of text. To shorten the procedure, we restrict its application to only those posts having a meaningful number of occurrences of the candidate name in the comments, as detailed in Section~\ref{sec:exp-res}.

\section{Experiments and Results}\label{sec:exp-res}

Here, we actualise the methodology in~\ref{subsec:meth} for the Facebook dataset described in Section~\ref{sec:datasets}.
In that dataset, the data reflect our hypothesised structure of ``post \& comments'': a post published onto a platform that accepts un-moderated (or, at least, not heavily moderated) comments.

Clearly, the original posts in the dataset are not censored: the names have not been altered, it being a condition that we need in order to check for the correctness of our de-censorship system. However, in order for us to work in a simulated censored condition, we need to first pre-process the dataset, by removing specific target names occurring in the source post. In order to do this, we rely on a generic Name Entity Recogniser to detect the names, which we then replace by unique random strings, to give the following processing steps no hints as to who the removed names were.

Therefore, to run the experiments according to the methodology defined in Section~\ref{subsec:meth},  we implement the steps listed below:
\begin{enumerate}
    \item selecting a target name that is the object of censorship in a Facebook post. We choose the name among a set of {\it popular} ones, specifically selected for the experiment, for which we expect to find enough data (Section \ref{subsec:selnames});
    \item retrieving a sub-set of the posts containing the target name, which will constitute the corpus that will be subject of censorship (Section \ref{subsec:retrieval});
    \item censoring such posts: the target name is removed from that sub-set of posts and it is replaced by a random string (pre-processing phase, Section \ref{subsec:retrieval});
    \item applying a NER to the comments tied to the sub-set of posts, to extract candidate names (Section \ref{subsec:comments}); 
    \item filtering the candidate names, so that only $k$ candidates remain (Section \ref{subsec:candidates});
    \item searching for the candidate names in the whole collection -- all the posts and comments in the dataset described in Section~\ref{sec:datasets} --  except for the set of posts from which the target name has been removed (Section \ref{subsection:examples}); \label{item:candidates}
    \item training a specific Candidate Entity Recogniser (CER), so that, instead of a generic ``person'' class, it is able to distinguish occurrences of the various candidates. The CER is trained on the data retrieved at step \ref{item:candidates} (Section \ref{subsection:training}); 
    \item applying the CER for the $k$ candidates to the censored name in the original set of posts, to see whether the classifier is able to correctly recognise the name that was censored (Section \ref{subsection:performance});
    \item evaluating the results in terms of standard measures, by comparing the decisions taken by the system {\it vs} the name actually removed from the posts (Section \ref{subsection:performance}).
\end{enumerate}

Below, we provide details of the implementation of each step.

\subsection{Selecting names}\label{subsec:selnames}
Names to test the system against were selected relying on a {\it popularity} criterion. In fact, we need names that are 1) present in the collection and 2) popular enough to appear in a certain amount of posts and comments. The latter requirement might appear as a strict constraint, but on Social Media data is often redundant.

To obtain the list of such names, we run a pre-trained NER over the whole data collection, retrieving all occurrences of person names. For this task, we use the spaCy\footnote{\url{https://spacy.io/docs/usage/entity-recognition}} Named Entity Recogniser in its pre-trained form. We make this particular choice because we need a NER that comes in pre-trained form, to ease our task, and that features the capability of being re-trained easily over new examples, for our purposes.
SpaCy is an open source software library for the Python language, entirely devoted to natural language processing. It includes several pre-trained models. The spaCy website reports an F-score for the used English NER of 85.30\footnote{\url{https://spacy.io/models/en}}.

The result of the NER application is a list of more than 1,000 terms, with their occurrences in the whole data collection.  We first polish the list, by removing terms erroneously recognised by the NER as person names (examples are ``Facebook", ``Wikileaks", ``Twitter", and ``Google"). Then, we keep only those names occurring more than 
100 times.
The result, consisting of 149 names, are listed in Table~\ref{tab:Terms-Politicians-full} and Table~\ref{tab:Terms-Celebrities-full}. 
For readability, we show here just a short excerpt of the two tables (Table~\ref{tab:Terms-Celebritiessummary}) and move the whole collection in \ref{appendix-names}.
Politicians turn out to represent the most present category in the collection (64 names). The other names, which mostly include journalists, entrepreneurs, activists, lawyers, athletes, TV stars and actors, are grouped together under the generic label ``Celebrities" (85 names).

\begin{table}[!ht]
\begin{centering}
\begin{footnotesize}
    \begin{tabular}{||c c||}
    \hline
    {\bf Politicians} & {\bf Freq}\\
    \hline
    Hillary Clinton & 23615 \\
    \hline
    Donald Trump & 17913 \\
    \hline
    Bill Clinton & 6668 \\
    \hline
    Gary Johnson & 3153 \\
    \hline
    Michelle Obama & 1079 \\
     \hline
    John McCain & 1079 \\
     \hline
    Bernie Sanders & 890 \\
     \hline
    Paul Ryan & 863 \\
     \hline
    Mike Pence & 859 \\
     \hline
    Barack Obama & 782 \\
     \hline
    Mitt Romney & 328 \\
        \hline
    John Podesta & 323  \\
        \hline
    Huma Abedin & 320 \\
        \hline
    Sarah Palin & 303 \\
         \hline
    Rudy Giuliani & 281 \\
         \hline
    Rick Scott &  248      \\
    \hline
    \end{tabular}
    \enskip
    \begin{tabular}{||c c||}
    \hline
    {\bf Celebrities} & {\bf Freq}\\
   \hline
    George Soros & 1173       \\
          \hline
    Margaret Sanger & 600     \\
          \hline
    James Comey & 585         \\
        \hline                 
    Paula Jones &  566        \\
        \hline                                             
    Billy Bush &  554         \\
        \hline                 
    Monica Levinsky &  438    \\
        \hline                 
    Colin Kaepernick &  401   \\
        \hline                 
    Julian Assange &  373     \\
        \hline
    Melania Trump &  362      \\
        \hline                 
    Saul Alinsky &  361       \\
        \hline                 
    Ryan Lochte &  320        \\
        \hline       
    Steve Bannon &  321       \\
        \hline                 
    Bill Cosby &  314         \\
        \hline                 
    Seymour Hersh &  304      \\
        \hline                 
    Ben Carson &  283         \\
        \hline                 
    John Kass &  280          \\
    \hline
    \end{tabular}
\end{footnotesize}
    \caption{Excerpt of politicians and celebrities with more than 100 occurrences.}
    \label{tab:Terms-Celebritiessummary}
    \end{centering}
\end{table}


\subsection{Retrieving and censoring the posts with the target names}\label{subsec:retrieval}
To retrieve all posts 
containing a certain name, we indexed the whole collection using Apache SOLR indexing and search engine\footnote{\url{http://lucene.apache.org/solr/}}. SOLR is a very popular keyword-based search engine system. Its core functionality is to index a text, and retrieve it by keywords (though more sophisticated means of retrieval are available). Its task is essentially defined as ``retrieve all and only the documents matching the query". The way we used in this paper, SOLR is responsible for:
\begin{enumerate}
    \item retrieving all documents containing a string (it is reasonable to expect that any bugless retrieval system performs this task with a 100\% accuracy);
    \item returning the window of text around the search string (same consideration as above).
\end{enumerate}
This was an enabling means for us to effectively retrieve all documents containing a name and then apply our methodology -- SOLR has no notion of the meaning of that name, so searching for ``John Wick" might return documents referring to the movie, or documents referring to a person by that name. Should SOLR be replaced by any alternative (ElasticSearch\footnote{\url{https://www.elastic.co/}}, or MySQL\footnote{\url{https://www.mysql.com/}}, or even \texttt{grep}'ping files), no changes to the algorithm or its accuracy would be observed. 

Clearly, searching for the occurrence of a name (e.g., ``Donald J. Trump'') does not guarantee that all the references to the same person are returned (a person might be referenced by aliases, nicknames, etc.), but, in its simplicity, it provides the highest possible precision, at the expense of recall, and it makes little difference for our purposes. Indeed, we will only consider those texts that are, in fact, returned.
Furthermore, using SOLR actually accomplishes a second feat to us. Since we will build a ``custom'' NER (i.e., our CER -- Candidate Entity Recogniser) on the immediate surrounding of names, we do not need the posts with the name in their whole form -- but just a window of text surrounding the occurrences of the name. Therefore, we can use SOLR's {\em snippet} feature to immediately retrieve a chunk of a few words surrounding the name instance. We asked for snippets of 200 characters and ignored all the snippets shorter than 50 characters. The choice of this particular length is due to the fact that it is coherent, e.g., with the length of a tweet or a Facebook comment. From these snippets,  we removed the actual name, replacing it with a token string composed of random letters (all unique to the snippet), with the first letter capitalised. This simulates our censorship of the snippets. These censored snippets are those bits of texts out of which we will try to reconstruct the original name.

It is worth noting that the target names are not replaced in the whole corpus of posts. Instead, we set as 20 the maximum number of posts where the name is replaced. In many cases, the name was present in less than 20 posts. The threshold has been chosen after diverse attempts, and 20 is the one that guarantees the best performances of our approach. 

\subsection{Retrieving candidates from comments}\label{subsec:comments}

For each SOLR resulting document censored so far, we retrieve the relative comments by exploiting the association between posts and their comments. In the comments, as per our assumption, we expect to find many names, among which the one we removed earlier from the snippet. To obtain a list of all names in the comments, we run the spaCy NER on them. The application of the NER to the comments produces, as a result, a list of names that we consider suitable candidates to fill the spot of the original name previously removed from the text snippet.


\subsection{Filtering candidates}\label{subsec:candidates}
An initial test showed that the candidates retrieved in \ref{subsec:comments} were often too many. Since we are going to train a specific Candidate Entity Recogniser (CER) to recognise them, we need to produce training examples for each of them, possibly a lengthy task. Therefore, we select only $k$ candidates from the list of all those returned at the previous step. The selection criteria is simple: we select the $k$ most frequent candidates found in the comments. It is worth noting that considering the  $k$ most frequent candidates might not include the actual name we are looking for. Thus, 
we always include the actual name within the $k$ filtered candidates. We remark that, even if we include the actual name in the $k$ candidates 
regardless of its frequency  among the comments, we verified that the name actually appears in the same comments. This preserves the fairness of the approach.

This also gives us a convenient baseline to compare our system against: what if the actual name is always the first one in the list of candidates? We would have solved the problem at this stage without further ado. In Section~\ref{subsection:performance}, we will compare the performance of our system against this na\"ive baseline solution.

\subsection{Fetching examples for each candidate}\label{subsection:examples}
After filtering the list of candidates, we need to figure out the typical context in which each of the candidates occurs. Clearly, the mentions in the comments tied to the censored post are a starting point, but we can use more examples. Still using SOLR, we tap into our dataset, searching for each candidate name and retrieving all the snippets in which it appears, them being relative to both posts and comments of the whole collection (excluding 
the original posts that we retrieved and censored in Section~\ref{subsec:retrieval}). It is worth noting that we make no attempt at reconciling the names. Thus, there is the possibility to obtain different sets of examples for 2 different names that might actually refer to the same person (e.g., ``Donald J Trump'' and ``The Donald''). This might have the disadvantage of spreading our training set that becomes too thin. In fact, still considering ``Donald J Trump'' and ``The Donald'', we could have two sets, one for ``Donald J Trump'' and the other for ``The Donald''; furthermore, obviously, the two sets would be smaller than their union.
However, it is of paramount importance to act in this way, to avoid the infusion in the system of a-priori knowledge beyond the data. It could also be the case that, when 2 (or more) names refer to the same person, the corresponding CER models will be very similar.
%
The fetched snippets of text constitute the training set for training our CER,  to recognise the name based on the surrounding text. For the purposes of the training, we only keep longish snippets ($>50$ chars).

\subsection{Training the Candidate Entity Recogniser}\label{subsection:training}
A Named Entity Recogniser usually works on broad entity categories, such as people, organisations, locations, etc. Being based on machine learning techniques, however, nothing keeps us from training a recogniser for a more specific task: the identification of not just any person, but, specifically, one of the $k$ candidates. To do that, we censor the snippets retrieved as in Section~\ref{subsection:examples} the same way we censored the original post, so that the CER is forced to build its model without relying on the specific occurrence of a name. In fact, usually, a NER would leverage features of the name itself (e.g., recognising the first given name). 
By removing the candidates' names, we force the model to rely on other features, i.e., the characteristics of the surrounding words. In order to be able to pinpoint a specific name, we annotate the censored names by one of the following $k$ classes: ANON for the occurrences of the target name, and DUMBO1, ..., DUMBO$k-1$ for the other candidates. Table~\ref{tab:nertrain} shows the way a NER is usually trained and the specific training we decided to implement here. 

\begin{table}[!ht]
\begin{small}
\begin{center}
    \begin{tabular} {||p{1.8in} p{0.2in} p{2.5in}||} 
     \hline
     \vspace{.75in}
     ``Standard'' NER annotation. &  &\texttt{As anyone with access to the internet can attest to, there hasn't exactly been a lack of
         news to pick from: President {\textless}PERSON{\textgreater}Donald Trump{\textless}/PERSON{\textgreater}'s possible collusion with the
         {\textless}ORGANISATION{\textgreater}Russian Kremlin{\textless}/ORGANISATION{\textgreater} during the 2016 election and Secretary of State {\textless}PERSON{\textgreater}Rex
         Tillerson{\textless}/PERSON{\textgreater}'s decision to break 18 years of tradition by declining to host a {\textless}EVENT{\textgreater}Ramadan{\textless}/EVENT{\textgreater}
         event at the state department are two prime examples.}
         \\
     \hline
    The NER annotation we used: in the example, Donald Trump is our name to find, Rex Tillerson is one of the candidates. Both have been replaced by random letters. For our purposes, organisations and events are ignored. & & \texttt{As anyone with access to the internet can attest to, there hasn't exactly been a lack of news to pick from: President {\textless}ANON{\textgreater}Xhyclertd{\textless}/ANON{\textgreater}'s possible collusion with the
         Russian Kremlin during the 2016 election and Secretary of State {\textless}DUMBO1{\textgreater}Vlargdiun{\textless}/DUMBO1{\textgreater}'s decision to break 18 years of tradition by declining to host a Ramadan
         event at the state department are two prime examples.} \\
     \hline
    \end{tabular}
\end{center}
\end{small}
 \caption{How a NER is normally trained {\it vs} how we train the CER for our purposes.}
\label{tab:nertrain}
\end{table}

\subsection{Resolving the target name}\label{subsection:performance}
The last step for finding out who the target name is applies the trained CER classifier to the censored snippets of Section~\ref{subsec:retrieval}. Given the way  the CER was built, we can check whether, in correspondence to a censored snippet, the CER returns the class ANON. This is indeed the class label that we assigned to the occurrences of the actual name removed from the snippet, and therefore, the correct classifier answer. All DUMBO$x$'s and, possibly, empty class assignments are to be considered wrong answers.

\subsection{Measuring performances}\label{subsection:performancemeasurement}

In addition to this, we can measure how hard the task is, by comparing our performances with the ones of two simple baselines: 1) the first baseline assigns to the censored snippet the most frequent candidate that appears in the related comments; 2) the second one assigns a choice at random among our $k$ candidates. Intuition might suggest that the first baseline could perform well, whereas the second one effectively represents a performance lower bound.
All the experiments were conducted with $k=5$, $10$, and $20$.

The CER is tested  over the names in Tables~\ref{tab:Terms-Politicians-full} and~\ref{tab:Terms-Celebrities-full}. In particular, over a number of 
149 names, occurring at least 100 times 
in the whole Facebook dataset, 
we consider only those with at least 50 occurrences in the comments related to the post where the name was censored, resulting in 
95 names. The 95 names are reported in Table~\ref{tab:95-Terms}  in the Appendix. 

For the evaluation of the system performances, we consider the following metrics. Given that the censored name is always inserted among the $k$ candidates (with $k$ = 5, 10, 20):
\begin{itemize}
    \item  The \emph{CER} accuracy is defined as the joint probability that 1) the system successfully recognises the censored name, and 2) the name is really one of the $k$ most frequent names in the comments associated to the censored posts.
    \item The \emph{Global} accuracy is the probability of the system to successfully recognise the censored name, regardless of the fact that the name is among the $k$ most frequent candidates (we remind the reader that the name is however present in the comments associated to the censored posts, thus guaranteeing the fairness of the approach).
    \item The \emph{Most frequent selection} accuracy is the probability that the most frequent candidate is the censored name (first baseline).
    \item  The \emph{Random among top $k$} accuracy is the probability to recognise the  censored name by randomly choosing from the top $k$ candidates (and being the name among such candidates, second baseline).
\end{itemize}

Remarkably, the CER accuracy gives the accuracy of the system when the analysis is performed choosing among the actual top $k$ most frequent names, over the associated comments per censored post. 
Thus, the CER accuracy and the Global accuracy match when the name is actually among the $k$ most frequent names in all the comments associated to the censored posts. Since the CER accuracy is computed as a joint probability, it holds that CER accuracy $\le$ Global accuracy $\le$ 1.

 \begin{table*}[!ht]
\begin{center}
 \begin{tabular}{|r|c|c|c|c|c|} 
 \hline 
 {\bf Target name} 
 & {\bf Post} 
 &\multicolumn{1}{|p{1.75cm}|}{\bf \centering CER \\ accuracy}
 &\multicolumn{1}{|p{1.75cm}|}{\bf \centering Global \\ accuracy}
 &\multicolumn{1}{|p{1.75cm}|}{\bf \centering Most \\ freq. \\ selection}
 &\multicolumn{1}{|p{1.75cm}|}{\bf \centering Random \\ among \\ top 10} \\

\hline
    Mitt Romney         & 10  & 0.20    & 0.40  & 0.00 & 0.07\\
    Rudy Giuliani       & 18  & 0.44    & 0.44  & 0.22 & 0.10\\
    Bernie Sanders      & 20  & 0.45    & 0.45  & 0.05 & 0.09\\
    Gary Johnson        & 20  & 0.45    & 0.45  & 0.35 & 0.10\\
    Mike Pence          & 20  & 0.50    & 0.50  & 0.30 & 0.10\\
    Rahm Emanuel        & 20  & 0.75    & 0.75  & 0.20 & 0.10\\
    Ryan Lochte         & 14  & 0.79    & 0.79  & 0.36 & 0.10\\
    Colin Kaepernick    & 20  & 0.75    & 0.85  & 0.45 & 0.09\\
    Paul Ryan           & 14  & 1.00    & 1.00  & 0.00 & 0.10\\
    Rick Scott          & 15  & 1.00    & 1.00  & 0.27 & 0.10\\
\hline\hline
    {\bf $\mu$avg (all 49 candidates)} & 10.94 & 0.43    & 0.55  & 0.14 & 0.07\\
\hline
\end{tabular}
\end{center}
    \caption{System performances: The worst 5 and top 5 results, considering target names censored in at least 10 posts. Settings: $k$ = 10, $nocc$ $\ge$ 200.}
    \label{tab:results}
\end{table*}
Table~\ref{tab:results} and Table~\ref{tab:summary} 
report the results under the following settings: $k$ = 10 and $nocc$ $\ge$  200 (where $nocc$ is the number of occurrences of the name in the whole collection).  Considering only the names that occur at least 200 times in the whole data collection and, among them, the ones that appear at least 50 times in the comments related to the post where those names was censored, we get a total of 49 names. Obviously, the 49 names are included in the 95 mentioned at the beginning of this Subsection~\ref{subsection:performance}. 
The complete outcome over the 49 names is in the Appendix, where Table~\ref{tab:results_complete_200}
shows the average results, both in terms of the single names, considering the number of posts in which the name has been censored, and in terms of the $\mu$average along all the candidates, considering the scenario with $k$ = 10. The $\mu$average is computed as the average of the single averages, per name. 

Table~\ref{tab:results} shows an excerpt of Table~\ref{tab:results_complete_200} in the Appendix. In particular, it reports the worst and best 5 results (in terms of Global accuracy) considering those target names censored in at least 10 posts. 
As an example, 
let the reader consider the case of ``Colin Kaepernick". Over the 20 posts in which the name was censored, the classifier correctly recognised the term in 75\% of the time, if the analysis is run considering the real 10 most frequent candidates per post. The accuracy rises to 85\% if we force to 1 the probability that the target name is among the 10 most frequent candidates, for the whole set of comments associated to each censored post.



Table~\ref{tab:summary} gives the overall statistics of the system, still evaluated over 49 names and $k$=10. It does not consider the single identities and it reports the flat average accuracy. Over a total number of 525 analysed posts, 49 censored names, and 10 possible candidates to choose from, the Candidate Entity Recogniser was able to correctly recognise the name 54\% of the time. When not considering the actual most frequent candidates per post, the average system accuracy rises to 0.62. Such results outperform the outcome of the two baselines. Notably, and probably not so intuitively, picking up the most frequent candidate mentioned in the comments as the censored name is successful only in 19\% of the cases. Even worse, choosing randomly among the 10 most frequent candidates leads to a success rate of about 10\%, as is to be expected. This is an indication of how hard the task is, and whereas 60\% performance might seem low for a classifier, the complexity of the task, and the simplifying assumptions must be taken into account. Regarding possible steps to improve the actual performances, we argue that the most direct direction to look into is widening the text window taken into account:  this was not done in this paper,  mainly because it further raises the issue of determining whether the enlarged window is relevant with respect to the censored name. Naively, here we assumed as relevant a short window around the occurrence.

\begin{table}[!ht]
\begin{center}
 \begin{tabular}{|r|c|c|c|c|} 
 \hline 
 {\bf Metric} & {\bf Value}\\
 \hline
    Total posts                     & 525   \\
    Total names                     & 49    \\
    Posts/names                     & 10.93 \\
    CER accuracy                    & 0.54  \\
    Global accuracy                 & 0.62  \\
    Most freq. selection accuracy   & 0.19  \\
    Random among top ten            & 0.09  \\
    \hline
\end{tabular}
\end{center}
    \caption{Overall statistics ($k$ = 10, $nocc$ $\ge$ 200).}
    \label{tab:summary}
\end{table}

Closing the discussion with settings $k$ = 10 and $nocc$ $\ge$ 200,  the classifier succeeds, on average, largely more than 50\% of the time, choosing among 10 different candidate names. We argue that the performance of our snippet classification approach is very promising.
 Indeed, it is worth noting how we heavily constrained our operational setting,  by considering concise snippets (50-200 characters each), both for posts and for comments. Furthermore, not all the comments related to a post are strictly related to the content of that post. Remarkably, the Facebook pages in which the posts are published contain the link to the complete news: we expect that considering the complete news text will lead to a sensitive improvement of the results. 
Finally, we notice that the length of our snippets is comparable to that of tweets, leading to the feasibility of applying the proposed approach over social platforms other than Facebook (e.g., Twitter and Tumblr).

\subsection{Performances of the classifier under different settings\label{subsec:diffscenarios}}
\begin{table}

    \centering
    \begin{scriptsize}
    \begin{tabular}{|r|c|c|c|c|c|}
        \hline
        \multicolumn{1}{|p{0,5cm}|} {\bf \textit{k}} 
        &\multicolumn{1}{|p{1,5cm}|} {\bf \textit{nocc}}
        &\multicolumn{1}{|p{2,5cm}|} {\bf Most freq. selection}
        &\multicolumn{1}{|p{2.5cm}|} {\bf Most freq. selection $\mu$ avg}
        &\multicolumn{1}{|p{2.5cm}|} {\bf Random among top $k$} &\multicolumn{1}{|p{2.5cm}|} {\bf Random among top $k$ $\mu$ avg}
        \\ \hline
        5   & $>100$ & 0.17 & 0.15 & 0.13 & 0.11 \\
        10  & $>100$ & 0.19 & 0.14 & 0.08 & 0.07 \\
        20  & $>100$ & 0.15 & 0.13 & 0.04 & 0.04 \\
        5  & $>200$ & 0.16 & 0.13 & 0.13 & 0.11 \\
        10  & $>200$ & 0.19 & 0.14 & 0.09 & 0.07 \\
        20  & $>200$ & 0.14 & 0.11 & 0.04 & 0.04\\
        \hline
    \end{tabular}

    \vspace{1mm}
    
    \begin{tabular}{|r|c|c|c|c|c|}
        \hline
        \multicolumn{1}{|p{0,5cm}|} {\bf \textit{k}} 
        &\multicolumn{1}{|p{1,5cm}|} {\bf \textit{nocc}}
        &\multicolumn{1}{|p{2,5cm}|} {\bf Global acc.} 
        &\multicolumn{1}{|p{2,5cm}|} {\bf Global acc. $\mu$ avg} 
        &\multicolumn{1}{|p{2,5cm}|}{\bf CER acc.} 
        &\multicolumn{1}{|p{2,5cm}|} {\bf CER acc. $\mu$ avg} \\
        \hline
        5   & $>100$ 
        & 0.61 & 0.49 & 0.39 & 0.26 \\
        10  & $>100$ 
        & 0.57 & 0.47 & 0.48 & 0.35 \\
        20  & $>100$ 
        & 0.48 & 0.39 & 0.39 & 0.29\\
        5  & $>200$ 
        & 0.65 & 0.57 & 0.42 & 0.31 \\
        10  & $>200$ 
        & 0.62 & 0.56 & 0.54 & 0.44 \\
        20  & $>200$ 
        & 0.53 & 0.49 & 0.43 & 0.34\\
        \hline
    \end{tabular}
    \end{scriptsize}

    \caption{System performances varying the number of candidates $k$ and the number of occurrences of the names $nocc$\label{tab:ALLsettings}.}
\end{table}
We conclude the presentation of the results by considering all the different settings in which we ran the experiments. Table~\ref{tab:ALLsettings} shows the system performances varying the values of $k$ and $nocc$. Focusing on the CER accuracy, the best average performances are achieved when considering the pair ($k$=10, $nocc$ $\ge$ 200): such results have been already presented and discussed in the previous section. Turning to the global accuracy, we achieve a slightly better result still considering only the names that appear at least 200 times in the whole collection, but having only 5 names as possible candidates (Global Accuracy = 0.65). 

Considering the number of occurrences of the target name in the whole collection, from the analysis of the values in the table, we can see that the worst performances are achieved with $nocc$ $\ge$ 100 (see, e.g., the column that reports the CER accuracy values, with 0.39, 0.48, and 0.39 $\le$ 0.42, 0.54, and 0.43, respectively). 

Instead, considering the number  $k$ of candidates to search within, taking into accounts all the 95 names ($nocc$ $\ge$ 100, first three lines in the table), the CER accuracy is noticeably higher when searching among 10 possible candidates (0.48) than that obtained with $k$ = 5 and $k$ = 20 (where the classifier achieves the same CER accuracy of 0.39). A different result is obtained for the Global Accuracy: the less the value of $k$, the better the accuracy.

The above considerations still hold considering the last three lines of the table (the ones with $nocc$ $\ge$ 200). 

For all the possible pairs of $k$ and $nocc$, there is an evident degradation of the performances, both when choosing the most frequent name as the censored name and when randomly guessing the name among the $k$ candidates.

Finally, Table~\ref{tab:results_complete_100} shows the complete results over all the tested names, with $k$ = 10.

\begin{figure}[!ht]
    \centering
    \begin{subfigure}{0.45\textwidth}
        \includegraphics[width=\textwidth]{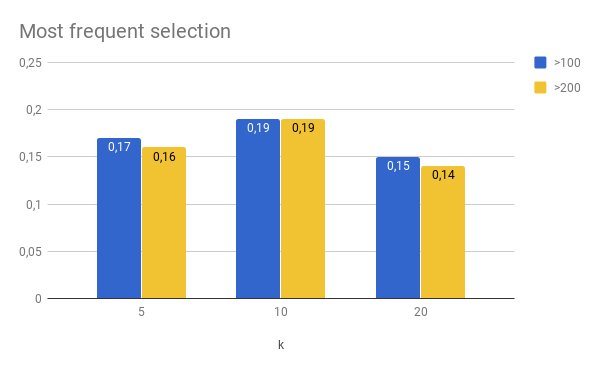}
        \caption{Probability that the censored name is the most frequent one}
        \label{fig:mostf-acc}
    \end{subfigure}
    ~
    \begin{subfigure}{0.45\textwidth}
        \includegraphics[width=\textwidth]{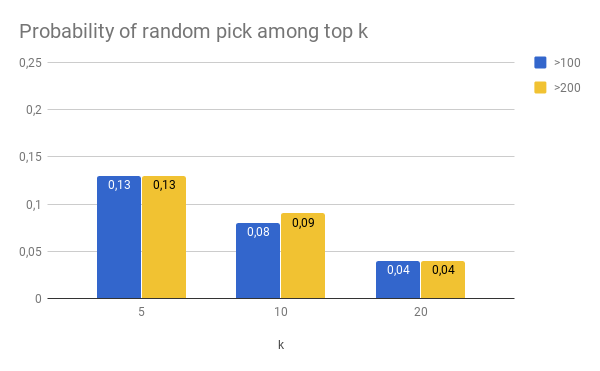}
        \caption{Probability of randomly selecting the censored name among the top $k$ candidates}
        \label{fig:rnd-acc}
    \end{subfigure}
    
    \begin{subfigure}{0.45\textwidth}
        \includegraphics[width=\textwidth]{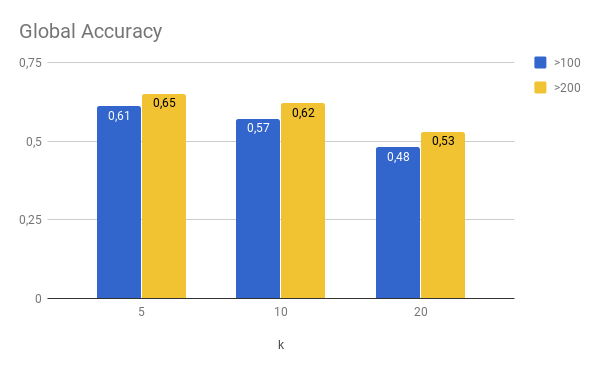}
        \caption{Global Accuracy}
        \label{fig:Glob-acc}
    \end{subfigure}
    ~
    \begin{subfigure}{0.45\textwidth}
        \includegraphics[width=\textwidth]{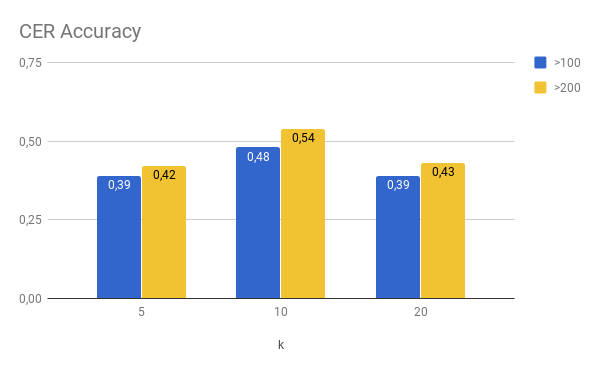}
        \caption{CER Accuracy}
        \label{fig:CER-acc}
    \end{subfigure}
    \caption{System performances at a glance: bar charts\label{fig:ALLsettings}.}
\end{figure}
Figure~\ref{fig:ALLsettings} shows in a pictorial way the classifier performances and those of the baseline techniques. Noticeably, the accuracy obtained when applying the baseline techniques (most frequent selection and random guessing among top $k$) are extremely poor, for all the tested combinations of $k$ and $nocc$ (Figures~\ref{fig:mostf-acc} and~\ref{fig:rnd-acc}). When applying the proposed classifier, considering only names that occur in the whole collection at least 200 times leads to better results than considering the names with at least 100 occurrences. This assumption holds independently from the value of $k$ for both Global and CER Accuracy (see the yellow bars in Figures~\ref{fig:Glob-acc} and~\ref{fig:CER-acc}, with respect to the blue bars in the same figures). 
Finally, as already noticed, the best results, in terms of CER Accuracy, are achieved with the configuration $k$=10, $nocc$ $\ge$ 200 (Figure~\ref{fig:CER-acc}). 
Overall, the Global Accuracy values are higher than the CER Accuracy values, since the former disregard the probability that the censored name is indeed in the top $k$ most frequent candidates (see the comparison between the values in Figure~\ref{fig:Glob-acc} and Figure~\ref{fig:CER-acc}).

\section{Related Work}\label{sec:relwork}
Social media provide Internet users with the opportunity to discuss, get
informed, express themselves and interact for a myriads of goals, such as planning events and engaging in commercial
transactions. In a word, users rely on online services to say to the
world what they are, think, do; and, viceversa, they learn the same
about the other subscribers. For a decade, scientists have been evaluating and assessing the attitude of users to disclose their personal information to receive higher exposure within the network community~\cite{Lindamood:2009,Lam2008}. 
Both sociologists  and computer scientists have investigated the amount and kind of personal information available on social networks. As an example, work in~\cite{Zhao2008} represents one of the first studies on identity construction on Facebook (comparing the difference in the identities narration on the popular social network with those on anonymous online environments). Two years later, the authors of~\cite{Nosko2010} studied the amount of personal information exposed by Facebook, characterising it according to the account age (the younger the user, the more the personal information exposed) and the inclination to set up new relationships. Thus, despite  the enormous volume of daily communications, which leads to levels of obfuscation and good resistance to analysis techniques~\cite{Conti:2016}, social networks naturally offer a huge amount of public  information -- even redundant -- with its fast diffusion supported by influencers and even weak ties among users~\cite{Bakshy2012,Guille2013}. Recently, researchers also concentrated on misinformation spread, considering the dynamics and motivations for the large number of followers of fake news~\cite{Bessi2015}. 
The demonstrated tendency of people to disclose their data, despite privacy issues~\cite{Min:2015}, has let researchers argue that, where the structure of data is under the form of a commented post, the content of comments may reveal a lot about the post itself, in those cases in which parts of them have been obfuscated. 
Indeed, several approaches have been tested in the literature to automatically classifying which comments lead to leakage of information. Some of these approaches exploit discourse analysis, to ``examine how language construct phenomena"~\cite{phillips2012} or semiotic analysis, which concerns the study of signs to infer the ``deeper meaning" of the data~\cite{grbich2012}. In~\cite{Schwartz:2017}, the authors shown how to leverage discourse and semiotic  analysis, in conjunction with standard text classification approaches, to automatically categorise leakage and non-leakage comments.


%

The issue of censorship on the Web has been faced from different perspectives: in the early 2000's, some information leakage was already possible to circumvent national censorship~\cite{Feamster2002,Feamster2003}.
In this work, we specifically consider censored texts. Work in~\cite{Serjantov2002} proposes a method to make textual documents resilient to censorship sharing them over a P2P network, one of the most frequently used decentralised sharing mechanism. However, when considering centralised systems, like Facebook, censorship might occur on document instances (i.e., posts and comments). Thus, strategic information may be altered - or censored-  by malicious users of P2P networks, as well as by authors of single posts on social media. 


 

Regarding news, work in~\cite{Schwartz:2017,Mascaro2012} characterises those comments exploitable for revealing censored data.
In~\cite{Cascavilla2015}, the authors consider comments to Facebook posts about Israeli military news. While specific identities in the post contents are censored, through the analysis of the social graph of the commenters~\cite{Burattin2014}, the authors were able to spot the identity of the mentioned people. In particular, the approach exploits accounts' public information to infer the ego network of  an individual and tries to reconstruct it when the access to user data is restricted from Facebook API, assuming that the account of the censored identity is linked to one of the commenters~\cite{Cascavilla:2016}. Thus, the approach is effective when the commenters are friends of the target, even in the case that comments are few. However, it might be less effective in general scenarios, in which commenters are not friends of the target of the censorship (like, e.g., when the censored identity is a popular one). Also, leveraging the social network structure, a significant a-priori knowledge is needed. 
The work in \cite{sharad2014} describes an effective approach to de-anonymize a social network using a random forest classifier and the number of friends of each node in the network as a feature. 

In a past work, the authors showed how to leverage a semi-supervised analysis approach to detects drugs and effects in large, domain-specific textual corpora~ \cite{Delvigna2016}.  Here, we inherit that snippets and contexts classification approach, to propose a methodology solely based on very loosely structured data. As clarified in the rest of the paper, we reveal identities in censored Facebook posts, only relying on a corpus made of very short texts, some of them even irrelevant to the domain of the post where the censored name is. This approach, although conceptually not far from what is proposed in \cite{Han2003,Han2005} for author disambiguation, adopts textual contents rather than other metadata to link information.
Disambiguation of identities on social networks is the matter of investigation in~\cite{Rowe2009,Rowe2010}, which exploits 
semantic social graphs to disambiguate among a set of possible person references.
The last approach is somehow in the middle between the de-anonymization technique proposed in~\cite{Burattin2014} and the semantic analysis performed in~\cite{Delvigna2016}, although the knowledge extraction is mainly performed for different purposes.

Compared with previous work with similar goals, the current proposal differentiates  because it does not rely on the social graph of the commenters to recognise the censored term. Instead, the paper proposes an adaptation - and application - of a text-analysis approach to the issue of unveiling censored identities. The approach is tested on a synthesised scenario, which however resembles a real use case. 

Furthermore, it is worth noting that, although a series of work consider automatic classification of comments to detect those ones leading to leakage of information, we decided here to bypass such a classification, and to consider the whole set of identities in the comments dataset. Thus, we ran a Name Entity Identifier to recognise the terms representing identities, and we directly pass to launch our methodology: this to distinguish, amongst the set of identified candidates, the censored term.


For the sake of completeness, we acknowledge the occurrence of different kinds of censorship, from DNS to router level ones~\cite{Verkamp2012}. Differently from veiling single terms, an entire domain might not be accessible, thus requiring different approaches to circumvent the block~\cite{Chen2010}. Work in~\cite{Gill2015, Leberknight2012part1, Leberknight2012part2} propose a survey on different topics related to censorship, either on detection or possible countermeasures. Also, monitoring tools exist, to trace the diffusion of the phenomenon~\cite{Nabi2013,Sfakianakis2011}.
Finally, emails or other communication channels different from social media might be affected by censorship~\cite{Nunziato2009}.

\section{Conclusions}\label{sec:concl}
In this paper, we applied a text analysis technique based on snippet classification to unveil censored identities in texts. As a running scenario, we considered the Facebook pages of the major US newspapers, considering posts to news and their related comments. The approach outperforms baseline techniques such as choosing randomly from a set of possible candidates or picking up the most frequent name mentioned in the comments.

 A limitation of the approach is given by the number of occurrences of the censored name that we need to find: in the experiments, we considered names 1) cited more than 100 times in the whole data collection, and 2) with at least 50 occurrences in the associated comments to the posts where the names appear. Also, the classifier performances moderately depend from the number of candidates names available. Among all the tested system configurations, we obtained the best results with a number of candidates to choose from equal to 10, and for names that occur at least 200 times in the whole collection. 
 
 Considering the average outcome of our classifier, in its best configuration, the accuracy of the system is largely above 50\%, meaning that we were able to identify the correct name in more than half of the tested cases (choosing among 10 possible candidates). This is an encouraging result. Indeed, we considered very short snippets (similar to tweets): on the one hand, this demonstrates the capability to apply the same technique to microblogging platforms like Twitter; on the other hand, this leaves room  for augmenting the performances, when considering longer texts (as an example for our scenario, the full version of the news, which is typically linked after the text of the post on the Facebook page). 
 
 Finally, it is worth noting that, due to its coverage, we considered Facebook as a relevant case study. However, our methodology is general enough to be applied to various data sources, provided there is a sufficient number of training examples. 
\clearpage

\appendix

\section{Further details on candidate names}\label{appendix-names}

\begin{table}[!ht]
\begin{centering}
\begin{footnotesize}
    \begin{tabular}{||c c||}
    \hline
    {\bf Politicians} & {\bf Freq}\\
    \hline
    Hillary Clinton & 23615 \\
    \hline
    Donald Trump & 17913 \\
    \hline
    Bill Clinton & 6668 \\
    \hline
    Gary Johnson & 3153 \\
    \hline
    Jill Stein & 1210 \\
    \hline
    Michelle Obama & 1079 \\
     \hline
    John McCain & 1079 \\
     \hline
    Bernie Sanders & 890 \\
     \hline
    Paul Ryan & 863 \\
     \hline
    Mike Pence & 859 \\
     \hline
    Barack Obama & 782 \\
     \hline
    Tim Kaine & 707 \\ 
      \hline
    Vladimir Putin & 686 \\
      \hline
    Al Gore & 529 \\
      \hline
    Harry Reid & 509 \\
      \hline
    George W. Bush & 419  \\
      \hline
    Ronald Reagan & 386 \\
       \hline
    David Duck & 379 \\ 
       \hline
    Bill Weld & 369 \\
       \hline
    Vince Foster & 361  \\
    \hline
    Colin Powell & 348 \\
       \hline
    Loretta Lynch & 335 \\
        \hline
    Antony Weiner & 333 \\
        \hline
    Elizabeth Warren  & 333 \\
        \hline
    Mitt Romney & 328 \\
        \hline
    John Podesta & 323  \\
        \hline
    Huma Abedin & 320 \\
        \hline
    Sarah Palin & 303 \\
         \hline
    Rudy Giuliani & 281 \\
         \hline
    Robert Byrd & 251 \\
         \hline
    Paul Manafort & 249 \\
         \hline
    Rick Scott & 248 \\
         \hline
    \end{tabular}
    \enskip
    \begin{tabular}{||c c||}
    \hline
    {\bf Politicians} & {\bf Freq}\\
    \hline
    Rahm Emanuel & 230 \\
    \hline 
    Debbie Wassermann & 228\\
          \hline
    Dick Cheney & 222\\
          \hline
    Chris Christie & 222 \\
          \hline
    Newt Gingrich & 216 \\
          \hline
    Marco Rubio & 215 \\
          \hline
    Joe Biden & 204 \\
    \hline
    Ken Starr & 192       \\
          \hline
    John Kerry & 188     \\
          \hline
    Sheriff Joe & 175         \\
        \hline                 
    Donna Brazile &  172        \\
        \hline                                             
    Roger Stone &  169         \\
        \hline                 
    Kellyanne Conway &  169    \\
        \hline                 
    Rand Paul &  168   \\
        \hline                 
    Richard Nixon &  166     \\
        \hline
    Jimmy Carter &  162      \\
        \hline      
        Kim Yong-un  &  162      \\
        \hline 
    Mitch McConnell &  158       \\
        \hline                 
    Pam Bondi &  151        \\
        \hline       
    Bashar al-Assad &  149       \\
        \hline                 
    Jeb Bush &  147         \\
        \hline                 
    Martin Luther King &  135      \\
        \hline                 
    John Brennan &  134         \\
        \hline                 
    Janet Reno &  133          \\
    \hline
    Jesse Jackson &  124          \\
        \hline                 
    Kelly Ayotte &  119  \\
        \hline                 
    George Duncan &  118    \\
        \hline                 
    Mark Kirk &  116     \\
        \hline                 
    Rob Portman &  116         \\
        \hline                 
    Thomas Jefferson &  113        \\
        \hline                 
    Bruce Rauner &  102      \\
        \hline                 
    Pat Toomey &  100         \\
        \hline                 
    \end{tabular}
\end{footnotesize}
    \caption{Politicians with more than 100 occurrences -- whole collection.}
    \label{tab:Terms-Politicians-full}
    \end{centering}
\end{table}

\begin{table}[!ht]
\begin{centering}
\begin{footnotesize}
    \begin{tabular}{||c c||}
    \hline
    {\bf Celebrities} & {\bf Freq}\\
   \hline
    George Soros & 1173       \\
          \hline
    Margaret Sanger & 600     \\
          \hline
    James Comey & 585         \\
        \hline                 
    Paula Jones &  566        \\
        \hline                                             
    Billy Bush &  554         \\
        \hline                 
    Monica Levinsky &  438    \\
        \hline                 
    Colin Kaepernick &  401   \\
        \hline                 
    Julian Assange &  373     \\
        \hline
    Melania Trump &  362      \\
        \hline                 
    Saul Alinsky &  361       \\
        \hline                 
    Ryan Lochte &  320        \\
        \hline       
    Steve Bannon &  321       \\
        \hline                 
    Bill Cosby &  314         \\
        \hline                 
    Seymour Hersh &  304      \\
        \hline                 
    Ben Carson &  283         \\
        \hline                 
    John Kass &  280          \\
    \hline
    Tom Brady &  278          \\
        \hline                 
    Juanita Broaddick &  272  \\
        \hline                 
    Andrew McCarthy &  260    \\
        \hline                 
    Michael Phelps &  259     \\
        \hline                 
    Shaun King &  257         \\
        \hline                 
    Lester Holt &  257        \\
        \hline                 
    Isabel Kilian &  253      \\
        \hline                 
    Mark Cuban &  235         \\
        \hline                 
    Frank Gaffney &  220      \\
        \hline                 
    Tony Shaffer &  218       \\
        \hline                 
    Rosie o'Donnell &  208    \\
        \hline                 
    Sean Hannity &  208       \\
        \hline                 
    Clair Lopez &  207        \\
        \hline                 
    Alex Jones &  205         \\
        \hline                 
    Megyn Kelly &  200        \\
        \hline
    Amy Schumer & 188       \\
          \hline
    Roger Ailes & 186     \\
          \hline
    Rupert Murdoch & 183         \\
        \hline                 
    Mark Westcott &  172        \\
        \hline                         
    Beyonce & 170         \\
        \hline                 
    Bill Murray &  168    \\
        \hline                 
    Jeffrey Epstein &  164   \\
        \hline                 
    Al Sharpton &  163     \\
        \hline
    Lillie Rose Fox &  162      \\
        \hline                 
    Alec Baldwin &  160       \\
        \hline                 
    Jerry Jones &  160        \\
        \hline       
    \end{tabular}
    \enskip
    \begin{tabular}{||c c||}
    \hline
    {\bf Celebrities} & {\bf Freq}\\
    \hline
     Sheldon Adelson &  157       \\
        \hline                 
    Madonna &  152         \\
        \hline                 
    Howard Stern &  152      \\
        \hline                 
    David Koresh &  151         \\
        \hline                 
    Ann Coulter &  149          \\
    \hline
    Anderson Cooper &  149          \\
        \hline                 
    Clint Eastwood &  148  \\
        \hline                 
    Matt Lauer &  144    \\
        \hline          
    John Hinckley &  143     \\
        \hline                 
    Gennifer Flowers &  143         \\
        \hline                 
    Ilene Jacobs &  142        \\
        \hline                 
    Warren Buffett &  139      \\
        \hline                 
    Gloria Allred &  137         \\
        \hline                 
    Bob Dylan &  135      \\
        \hline                 
    Rachel Flowers &  135       \\
        \hline                 
    Brandon Marshall &  133    \\
        \hline                 
    Zoe Baird &  132       \\
        \hline                 
    Chris Wallace &  131        \\
        \hline                 
    David Hinckley &  129         \\
        \hline                 
    Mike Evans &  128        \\
    \hline
    Branch Davidian &  125    \\
    \hline
    Garrison Keillor &  127    \\
        \hline                 
    William Kennedy &  126     \\
        \hline                 
    John Oliver &  126         \\
        \hline                 
    Billy Dale &  124        \\
        \hline                 
    Tony Romo &  123      \\
        \hline                 
    Brock Turner &  121         \\
        \hline                 
    Alicia Machado &  118      \\
        \hline                 
    Rachel Flowers &  135       \\
        \hline                 
    Khizr Khan  &  115    \\
        \hline                 
    Hope Solo &  114       \\
        \hline                 
    Michael Moore &  112        \\
        \hline                 
    Kim Kardashian &  110         \\
        \hline                 
   Michael Jackson &  108        \\
    \hline
    Rush Limbaugh &  107       \\
        \hline                 
    Brian Johnston  &  107    \\
        \hline                 
    Scott Baio &  106       \\
        \hline                 
    Chelsea Clinton &  105        \\
        \hline                 
    Pope Francis &  105         \\
        \hline                 
   Dan Rather &  104        \\
   \hline
   Ted Nugent   &  103    \\
        \hline                 
    Kevin Hart &  101       \\
        \hline                 
    Patrick Grant &  100        \\
        \hline                 
    \end{tabular}
\end{footnotesize}
    \caption{Celebrities with more than 100 occurrences -- whole collection.}
    \label{tab:Terms-Celebrities-full}
    \end{centering}
\end{table}

\begin{table}[!ht]
\begin{centering}
\begin{scriptsize}
    \begin{tabular}{||c c||}
    \hline
    {\bf Politicians} & {\bf Freq}\\
   \hline
    Hillary Clinton & 23615       \\
          \hline
    Donald Trump & 17913     \\
          \hline
    Bill Clinton & 6668         \\
        \hline   
        Gary Johnson & 3153 \\
        \hline
    Jill Stein &  1210       \\
        \hline                                         
    Michelle Obama &  1079         \\
        \hline                 
    Bernie Sanders & 890    \\
        \hline                 
    Paul Ryan &  863   \\
        \hline                 
    Mike Pence &  859     \\
        \hline
    Barack Obama &  782      \\
        \hline                 
    Tim Kaine &  707       \\
        \hline                 
    Vladimir Putin &  686        \\
        \hline       
    Al Gore &  529       \\
        \hline                 
    Harry Reid &  509         \\
        \hline                 
    George W. Bush &  419      \\
        \hline                 
    Ronald Reagan &  386         \\
        \hline                 
    Anthony Weiner &  333          \\
    \hline
    Elizabeth Warren &  333          \\
        \hline                 
    Mitt Romney &  328  \\
        \hline                 
    John Podesta &  323    \\
        \hline                 
    Huma Abedin &  320     \\
        \hline                 
    Rudy Giuliani &  281         \\
        \hline                 
    Paul Manafort &  249        \\
        \hline                 
    Rick Scott &  248      \\
        \hline 
        Rahm Emanuel & 230 \\
        \hline 
    Chris Christie &  222         \\
        \hline                 
    Newt Gingrich &  216      \\
        \hline                 
    Marco Rubio &  215       \\
        \hline                 
    Joe Biden &  204    \\
        \hline                 
    John Kerry &  188       \\
        \hline                 
    Sheriff Joe &  175        \\
        \hline                 
    Donna Brazile &  172         \\
        \hline                 
    Roger Stone &  169        \\
        \hline
    Kellyanne Conway & 169       \\
          \hline
    Richard Nixon & 166     \\
          \hline
    Jimmy Carter & 162\\
          \hline
    Kim Jong-un  & 162\\
          \hline
    Pam Bondi & 151         \\
        \hline                 
    Jeb Bush &  147        \\
        \hline                         
    Martin Luther King & 135         \\
        \hline                 
    Janet Reno &  133    \\
        \hline                 
    Jesse Jackson &  124   \\
        \hline                 
    Kelly Ayotte &  119     \\
        \hline
    Mark Kirk &  116      \\
        \hline                 
    Bruce Rauner &  102       \\
        \hline      
    \end{tabular}
    \enskip
    \begin{tabular}{||c c||}
    \hline
    {\bf Celebrities} & {\bf Freq}\\
    \hline
        George Soros & 1173       \\
          \hline
    James Comey & 585         \\
        \hline                 
    Paula Jones &  566        \\
        \hline    
    Billy Bush &  554         \\
        \hline                 
    Monica Levinsky &  438    \\
        \hline                 
    Colin Kaepernick &  401   \\
        \hline                 
    Julian Assange &  373     \\
        \hline
    Melania Trump &  362      \\
        \hline                 
    Ryan Lochte &  320        \\
        \hline       
    Steve Bannon &  321       \\
        \hline                 
    John Kass &  280          \\
    \hline
    Tom Brady &  278          \\
        \hline                 
    Juanita Broaddick &  272  \\
        \hline                 
    Michael Phelps &  259     \\
        \hline                 
    Shaun King &  257         \\
        \hline                 
    Lester Holt &  257        \\
        \hline                 
    Mark Cuban &  235         \\
        \hline                 
    Sean Hannity &  208       \\
        \hline                 
    Alex Jones &  205         \\
        \hline     
        Megyn Kelly & 200 \\
        \hline
    Amy Schumer & 188       \\
          \hline
    Roger Ailes & 186     \\
          \hline
    Beyonce & 170         \\
        \hline                 
    Bill Murray &  168    \\
        \hline                 
    Alec Baldwin &  160       \\
        \hline                 
    Jerry Jones &  160        \\
        \hline  
     Sheldon Adelson &  157       \\
        \hline                 
    Madonna &  152         \\
        \hline                 
    Howard Stern &  152      \\
        \hline                 
    Anderson Cooper &  149          \\
        \hline                 
    Clint Eastwood &  148  \\
        \hline                 
    Matt Lauer &  144    \\
        \hline           
    John Hinckley &  143     \\
        \hline                 
    Gennifer Flowers &  143         \\
        \hline                 
    Warren Buffett &  139      \\
        \hline                 
    Brandon Marshall &  133    \\
        \hline                 
    Chris Wallace &  131        \\
        \hline                 
    Mike Evans &  128        \\
    \hline
    Garrison Keillor &  127    \\
        \hline              
        Branch Davidian & 125 \\
        \hline
    John Oliver &  126         \\
        \hline                 
    Tony Romo &  123      \\
        \hline                 
    Brock Turner &  121         \\
        \hline                 
    Alicia Machado &  118      \\
        \hline                 
    Khizr Khan  &  115    \\
        \hline                 
    Hope Solo &  114       \\
        \hline                 
    Kim Kardashian &  110         \\
        \hline                 
    Chelsea Clinton &  105        \\
        \hline                 
    Pope Francis &  105         \\
        \hline                 
    Kevin Hart &  101       \\
        \hline                 
    \end{tabular}
\end{scriptsize}
    \caption{Candidates with at least 100 occurrences in the whole collection and at least 50 occurrences in the comments associated to the post where the name is censored}
    \label{tab:95-Terms}
    \end{centering}
\end{table}

\clearpage

\section{Classifier Performances -- $k$ = 10, $nocc$ $\ge$ 200}\label{appendix-200}

\begin{table}[!ht]
\begin{scriptsize}
\begin{center}
 \begin{tabular}{|r|c|c|c|c|c|} 
 \hline
 {\bf Target name} 
 & {\bf Post} 
 &\multicolumn{1}{|p{2cm}|}{\bf \centering CER \\ accuracy}
 &\multicolumn{1}{|p{2cm}|}{\bf \centering Global \\ accuracy}
 &\multicolumn{1}{|p{3cm}|}{\bf \centering Most freq. \\ selection  accuracy}
 &\multicolumn{1}{|p{2.5cm}|}{\bf \centering Random among\\top 10 accuracy} \\
 \hline
Al Gore		       & 3	&   0.00        &   0.00    &    0.00   &    0.07  \\ 
Alex Jones		   & 1	&   1.00        &   1.00    &    0.00   &    0.10  \\ 
Anthony Weiner	   & 18	&   0.50        &   0.56    &    0.11   &    0.09  \\ 
Barack Obama	   & 20	&   0.60        &   0.75    &    0.00   &    0.09  \\ 
Bernie Sanders	   & 20	&   0.45        &   0.45    &    0.05   &    0.09  \\ 
Bill Clinton	   & 20	&   0.55        &   0.55    &    0.00   &    0.10  \\ 
Billy Bush		   & 15	&   0.47        &   0.67    &    0.00   &    0.07  \\ 
Chris Christie	   & 7	&   0.14        &   0.43    &    0.00   &    0.01  \\ 
Colin Kaepernick   & 20	&   0.75        &   0.85    &    0.45   &    0.09  \\ 
Donald Trump	   & 20	&   0.65        &   0.65    &    1.00   &    0.10  \\ 
Elizabeth Warren   & 11	&   0.36        &   0.55    &    0.09   &    0.07  \\ 
Gary Johnson	   & 20	&   0.45        &   0.45    &    0.35   &    0.10  \\ 
George Bush		   & 15	&   0.40        &   0.60    &    0.13   &    0.07  \\ 
George Soros	   & 2	&   0.00        &   0.50    &    0.00   &    0.00  \\ 
Harry Reid		   & 12	&   0.58        &   0.75    &    0.17   &    0.08  \\ 
Hillary Clinton	   & 20	&   0.60        &   0.60    &    0.35   &    0.10  \\ 
Huma Abedin		   & 12	&   0.42        &   0.75    &    0.08   &    0.06  \\ 
James Comey		   & 20	&   0.60        &   0.75    &    0.00   &    0.08  \\ 
Jill Stein		   & 11	&   0.73        &   0.73    &    0.00   &    0.10  \\ 
Joe Biden		   & 7	&   0.57        &   1.00    &    0.14   &    0.06  \\ 
John Kass		   & 16	&   0.75        &   0.81    &    0.06   &    0.09  \\ 
John Podesta	   & 13	&   0.23        &   0.69    &    0.00   &    0.03  \\ 
Juanita Broaddrick & 1	&   0.00        &   0.00    &    0.00   &    0.00  \\ 
Julian Assange	   & 1	&   0.00        &   1.00    &    0.00   &    0.00  \\ 
Lester Holt		   & 2	&   0.00        &   0.00    &    0.00   &    0.05  \\ 
Marco Rubio		   & 9	&   0.22        &   0.33    &    0.44   &    0.07  \\ 
Mark Cuban		   & 6	&   0.00        &   0.17    &    0.00   &    0.05  \\ 
Megyn Kelly		   & 9	&   0.56        &   0.67    &    0.00   &    0.08  \\ 
Melania Trump	   & 20	&   0.60        &   0.60    &    0.80   &    0.10  \\ 
Michael Phelps	   & 1	&   0.00        &   1.00    &    0.00   &    0.00  \\ 
Michelle Obama	   & 20	&   0.50        &   0.50    &    0.05   &    0.10  \\ 
Mike Pence		   & 20	&   0.50        &   0.50    &    0.30   &    0.10  \\ 
Mitt Romney		   & 10	&   0.20        &   0.40    &    0.00   &    0.07  \\ 
Monica Levinski	   & 1	&   0.00        &   0.00    &    0.00   &    0.00  \\ 
Newt Gingrich	   & 8	&   0.25        &   0.38    &    0.25   &    0.06  \\ 
Paul Manafort	   & 1	&   0.00        &   0.00    &    0.00   &    0.10  \\ 
Paul Ryan		   & 14	&   1.00        &   1.00    &    0.00   &    0.10  \\ 
Paula Jones		   & 2	&   0.50        &   0.50    &    0.00   &    0.10  \\ 
Rahm Emanuel	   & 20	&   0.75        &   0.75    &    0.20   &    0.10  \\ 
Rick Scott		   & 15	&   1.00        &   1.00    &    0.27   &    0.10  \\ 
Ronald Regan	   & 4	&   0.50        &   0.50    &    0.25   &    0.10  \\ 
Rudy Giuliani	   & 18	&   0.44        &   0.44    &    0.22   &    0.10  \\ 
Ryan Lochte		   & 14	&   0.79        &   0.79    &    0.36   &    0.10  \\ 
Sean Hannity	   & 2	&   0.00        &   0.00    &    0.00   &    0.05  \\ 
Shaun King		   & 5	&   1.00        &   1.00    &    0.00   &    0.10  \\ 
Steve Bannon	   & 2	&   0.00        &   0.00    &    0.50   &    0.10  \\ 
Tim Kaine		   & 17	&   0.53        &   0.53    &    0.00   &    0.10  \\ 
Tom Brady		   & 7	&   0.71        &   0.71    &    0.14   &    0.10  \\ 
Vladimir Putin	   & 9	&   0.56        &   0.56    &    0.00   &    0.10  \\ 

\hline
{\bf $\mu$avg}     & 11.04 & 0.54       &   0.56    &    0.14   &    0.07\\
\hline
\end{tabular}
 \end{center}
\end{scriptsize}
    \caption{Performances for names with at least 200 occurrences\label{tab:results_complete_200}}
\end{table}
 
\clearpage

\section{Classifier Performances -- $k$ = 10, $nocc$$\ge$ 100}\label{appendix-100}

\begin{center} 
\begin{scriptsize}
\begin{longtable}{|r|c|c|c|c|c|}
 \hline
 {\bf Target name} 
 & {\bf Post} 
 &\multicolumn{1}{|p{2cm}|}{\bf \centering CER \\ accuracy}
 &\multicolumn{1}{|p{2cm}|}{\bf \centering Global \\ accuracy}
 &\multicolumn{1}{|p{3cm}|}{\bf \centering Most freq. \\ selection  accuracy}
 &\multicolumn{1}{|p{2.5cm}|}{\bf \centering Random among\\top 10 accuracy} \\

\hline
Al Gore		       &     3	& 0.00  &   0.00    &    0.00       &    0.07 \\  
Alec Baldwin	   &     5	& 0.00  &   0.00    &    0.00       &    0.08 \\  
Alex Jones		   &     1	& 1.00  &   1.00    &    0.00       &    0.10 \\  
Alicia Machado	   &     10	& 0.20  &   0.30    &    0.00       &    0.03 \\  
Amy Schumer		   &     14	& 0.29  &   0.29    &    0.07       &    0.10 \\  
Anderson Cooper	   &     3	& 0.00  &   0.67    &    0.00       &    0.00 \\  
Anthony Weiner	   &     18	& 0.50  &   0.56    &    0.11       &    0.09 \\  
Barack Obama	   &     20	& 0.60  &   0.75    &    0.00       &    0.09 \\  
Bernie Sanders	   &     20	& 0.45  &   0.45    &    0.05       &    0.09 \\  
Beyonce		       &     1	& 0.00  &   0.00    &    0.00       &    0.10 \\  
Bill Clinton	   &     20	& 0.55  &   0.55    &    0.00       &    0.10 \\  
Bill Murray		   &     4	& 0.25  &   0.25    &    0.50       &    0.10 \\  
Billy Bush		   &     15	& 0.47  &   0.67    &    0.00       &    0.07 \\  
Branch Davidian	   &     1	& 0.00  &   0.00    &    0.00       &    0.00 \\  
Brandon Marshall   &     6	& 0.17  &   0.17    &    0.33       &    0.08 \\  
Brock Turner	   &     4	& 0.00  &   0.50    &    0.00       &    0.05 \\  
Bruce Rauner	   &     8	& 0.75  &   0.75    &    0.00       &    0.10 \\  
Chelsea Clinton	   &     7	& 0.00  &   0.43    &    0.00       &    0.03 \\  
Chris Christie	   &     7	& 0.14  &   0.43    &    0.00       &    0.01 \\  
Chris Wallace	   &     1	& 1.00  &   1.00    &    0.00       &    0.10 \\  
Clint Eastwood	   &     1	& 0.00  &   0.00    &    0.00       &    0.10 \\  
Colin Kaepernick   &     20	& 0.75  &   0.85    &    0.45       &    0.09 \\  
Donald Trump	   &     20	& 0.65  &   0.65    &    1.00       &    0.10 \\  
Donna Brazile	   &     2	& 0.50  &   0.50    &    0.00       &    0.05 \\  
Elizabeth Warren   &     11	& 0.36  &   0.55    &    0.09       &    0.07 \\  
Garrison Keillor   &     3	& 0.33  &   0.33    &    0.00       &    0.10 \\  
Gary Johnson	   &     20	& 0.45  &   0.45    &    0.35       &    0.10 \\  
Gennifer Flowers   &     2	& 0.50  &   0.50    &    0.00       &    0.10 \\  
George Bush		   &     15	& 0.40  &   0.60    &    0.13       &    0.07 \\  
George Soros	   &     2	& 0.00  &   0.50    &    0.00       &    0.00 \\  
Harry Reid		   &     12	& 0.58  &   0.75    &    0.17       &    0.08 \\  
Hillary Clinton	   &     20	& 0.60  &   0.60    &    0.35       &    0.10 \\  
Hope Solo		   &     1	& 1.00  &   1.00    &    1.00       &    0.10 \\  
Howard Stern	   &     3	& 0.00  &   0.67    &    0.00       &    0.03 \\  
Huma Abedin		   &     12	& 0.42  &   0.75    &    0.08       &    0.06 \\  
James Comey		   &     20	& 0.60  &   0.75    &    0.00       &    0.08 \\  
Janet Reno		   &     1	& 1.00  &   1.00    &    0.00       &    0.10 \\  
Jeb Bush		   &     3	& 0.00  &   0.33    &    0.00       &    0.00 \\  
Jerry Jones		   &     3	& 0.00  &   0.00    &    0.00       &    0.10 \\  
Jesse Jackson	   &     3	& 0.67  &   0.67    &    0.00       &    0.10 \\  
Jill Stein		   &     11	& 0.73  &   0.73    &    0.00       &    0.10 \\  
Jimmy Carter	   &     1	& 0.00  &   0.00    &    0.00       &    0.00 \\  
Joe Biden		   &     7	& 0.57  &   1.00    &    0.14       &    0.06 \\  
John Hinckley	   &     1	& 0.00  &   0.00    &    0.00       &    0.10 \\  
John Kass		   &     16	& 0.75  &   0.81    &    0.06       &    0.09 \\  
John Kerry		   &     2	& 0.50  &   0.50    &    0.50       &    0.10 \\  
John Oliver		   &     1	& 0.00  &   0.00    &    0.00       &    0.10 \\  
John Podesta	   &     13	& 0.23  &   0.69    &    0.00       &    0.03 \\  
Juanita Broaddrick &     1	& 0.00  &   0.00    &    0.00       &    0.00 \\  
Julian Assange	   &     1	& 0.00  &   1.00    &    0.00       &    0.00 \\  
Kelly Ayotte	   &     7	& 0.29  &   0.29    &    0.14       &    0.09 \\  
Kellyanne Conway   &     1	& 0.00  &   0.00    &    0.00       &    0.00 \\  
Kevin Hart		   &     1	& 0.00  &   0.00    &    0.00       &    0.10 \\  
Khizr Khan		   &     6	& 0.17  &   0.17    &    0.00       &    0.10 \\  
Kim Jong-Un		   &     1	& 1.00  &   1.00    &    0.00       &    0.10 \\  
Kim Kardashian	   &     9	& 0.56  &   0.56    &    0.33       &    0.10 \\  
Lester Holt		   &     2	& 0.00  &   0.00    &    0.00       &    0.05 \\  
Madonna		       &     1	& 0.00  &   0.00    &    0.00       &    0.00 \\  
Marco Rubio		   &     9	& 0.22  &   0.33    &    0.44       &    0.07 \\  
Mark Cuban		   &     6	& 0.00  &   0.17    &    0.00       &    0.05 \\  
Mark Kirk		   &     8	& 0.75  &   0.88    &    0.63       &    0.09 \\  
Martin Luther King &     2	& 0.00  &   1.00    &    0.00       &    0.00 \\  
Matt Lauer		   &     2	& 0.00  &   0.00    &    0.00       &    0.10 \\  
Megyn Kelly		   &     9	& 0.56  &   0.67    &    0.00       &    0.08 \\  
Melania Trump	   &     20	& 0.60  &   0.60    &    0.80       &    0.10 \\  
Michael Phelps	   &     1	& 0.00  &   1.00    &    0.00       &    0.00 \\  
Michelle Obama	   &     20	& 0.50  &   0.50    &    0.05       &    0.10 \\  
Mike Evans		   &     6	& 0.33  &   0.33    &    0.67       &    0.10 \\  
Mike Pence		   &     20	& 0.50  &   0.50    &    0.30       &    0.10 \\  
Mitt Romney		   &     10	& 0.20  &   0.40    &    0.00       &    0.07 \\  
Monica Levinski	   &     1	& 0.00  &   0.00    &    0.00       &    0.00 \\  
Newt Gingrich	   &     8	& 0.25  &   0.38    &    0.25       &    0.06 \\  
Pam Bondi		   &     8	& 0.75  &   0.75    &    0.00       &    0.10 \\  
Paul Manafort	   &     1	& 0.00  &   0.00    &    0.00       &    0.10 \\  
Paul Ryan		   &     14	& 1.00  &   1.00    &    0.00       &    0.10 \\  
Paula Jones		   &     2	& 0.50  &   0.50    &    0.00       &    0.10 \\  
Pope Francis	   &     3	& 0.33  &   0.67    &    0.00       &    0.03 \\  
Rahm Emanuel	   &     20	& 0.75  &   0.75    &    0.20       &    0.10 \\  
Richard Nixon	   &     5	& 0.20  &   0.20    &    0.00       &    0.02 \\  
Rick Scott		   &     15 & 1.00  &   1.00    &    0.27       &    0.10 \\  
Roger Ailes		   &     2	& 0.00  &   0.00    &    0.00       &    0.05 \\  
Roger Stone		   &     2	& 0.00  &   0.50    &    0.00       &    0.00 \\  
Ronald Regan	   &     4	& 0.50  &   0.50    &    0.25       &    0.10 \\  
Rudy Giuliani	   &     18	& 0.44  &   0.44    &    0.22       &    0.10 \\  
Ryan Lochte		   &     14	& 0.79  &   0.79    &    0.36       &    0.10 \\  
Sean Hannity	   &     2	& 0.00  &   0.00    &    0.00       &    0.05 \\  
Shaun King		   &     5  & 1.00  &   1.00    &    0.00       &    0.10 \\  
Sheldon Adelson	   &     1	& 0.00  &   0.00    &    1.00       &    0.10 \\  
Sheriff Joe		   &     9	& 0.33  &   0.33    &    0.56       &    0.10 \\  
Steve Bannon	   &     2	& 0.00  &   0.00    &    0.50       &    0.10 \\  
Tim Kaine		   &     17	& 0.53  &   0.53    &    0.00       &    0.10 \\  
Tom Brady		   &     7	& 0.71  &   0.71    &    0.14       &    0.10 \\  
Tony Romo		   &     5	& 0.40  &   0.40    &    0.80       &    0.10 \\  
Vladimir Putin	   &     9	& 0.56  &   0.56    &    0.00       &    0.10 \\  
Warren Buffett	   &     3	& 0.00  &   0.00    &    0.00       &    0.03 \\  

\hline
{\bf $\mu$avg}     &     7.52 & 0.35 &  0.47    &    0.14       &    0.07\\
    \hline
    \caption{Performances over names with at least 100 occurrences\label{tab:results_complete_100}}
\end{longtable}
\end{scriptsize}
\end{center}

\clearpage




\bibliographystyle{elsarticle-num}

\section*{References}

\bibliography{bibliography}


\end{document}